\pdfoutput=1
\documentclass{article}
\usepackage{spconf,amsmath,graphicx}
\usepackage{stmaryrd}
\usepackage{amssymb}
\usepackage{caption}
\usepackage{hyperref}
\usepackage{xcolor}
\usepackage[symbol]{footmisc}
\usepackage{dblfloatfix}
\DeclareMathOperator*{\argmin}{arg\,min}

\makeatletter

\title{StreaMulT: Streaming Multimodal Transformer for Heterogeneous and Arbitrary Long Sequential Data}
\name{Victor Pellegrain$^{\dagger \star}$, Myriam Tami$^\star$, Michel Batteux$^\dagger$, Céline Hudelot$^\star$ }
\address{$^\dagger$ Institut de Recherche Technologique SystemX, 2 boulevard Thomas Gobert, 91120, Palaiseau, France \\ $^\star$ Université Paris-Saclay, CentraleSupélec, MICS, 91190, Gif-sur-Yvette, France}

\begin{document}
\ninept
\maketitle

\begin{abstract}
The increasing complexity of Industry 4.0 systems brings new challenges regarding predictive maintenance tasks such as fault detection and diagnosis. A corresponding and realistic setting includes multi-source data streams from different modalities, such as sensors measurements time series, machine images, textual maintenance reports, etc. These heterogeneous multimodal streams also differ in their acquisition frequency, may embed temporally unaligned information and can be arbitrarily long, depending on the considered system and task. Whereas multimodal fusion has been largely studied in a static setting, to the best of our knowledge, there exists no previous work considering arbitrarily long multimodal streams alongside with related tasks such as prediction across time. Thus, in this paper, we first formalize this paradigm of heterogeneous multimodal learning in a streaming setting as a new one. To tackle this challenge, we propose StreaMulT, a Streaming Multimodal Transformer relying on cross-modal attention and on a memory bank to process arbitrarily long input sequences at training time and run in a streaming way at inference. StreaMulT improves the state-of-the-art metrics on CMU-MOSEI dataset for Multimodal Sentiment Analysis task, while being able to deal with much longer inputs than other multimodal models. The conducted experiments eventually highlight the importance of the textual embedding layer, questioning recent improvements in Multimodal Sentiment Analysis benchmarks.
\end{abstract}
\begin{keywords}
Multimodal learning, Streaming data, Transformer, Long-term dependencies
\end{keywords}
\section{Introduction}
\label{sec:intro}

\begin{figure*}[h]
    \centering
    \includegraphics[width =\textwidth]{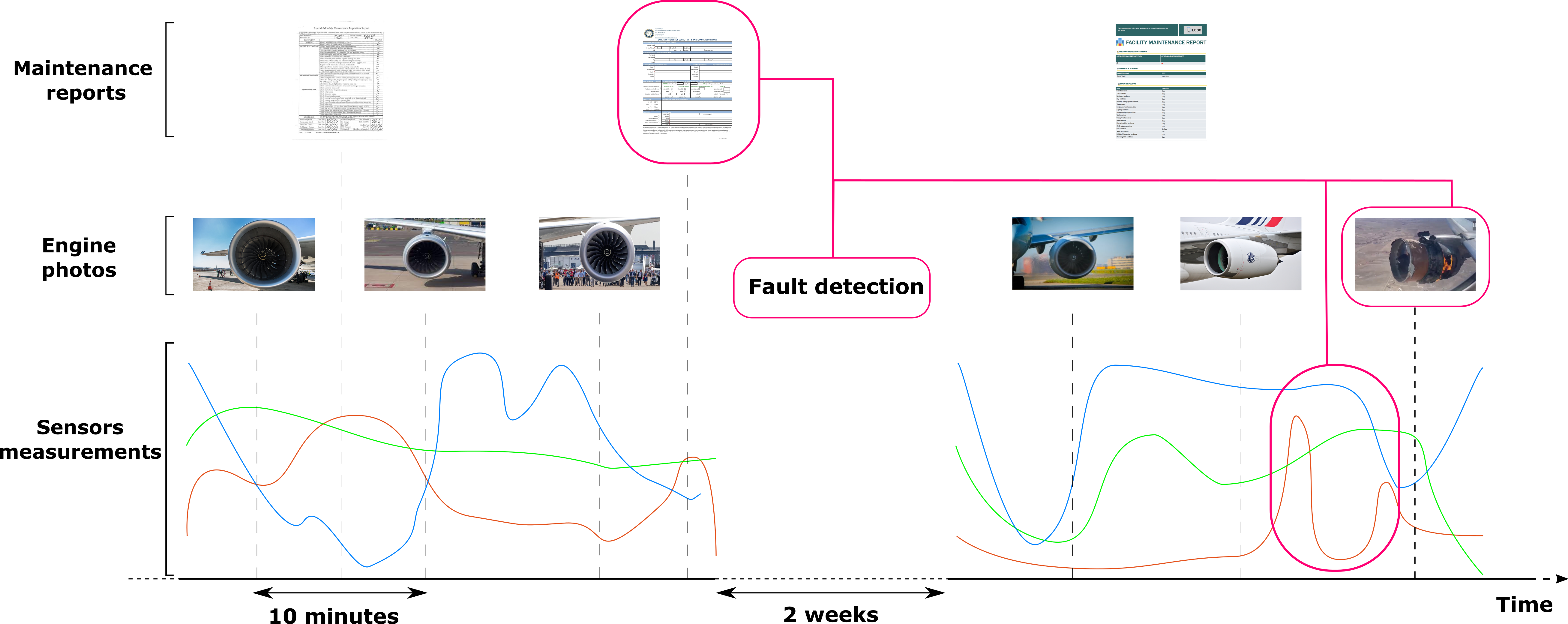}
    \caption{Multimodal learning in a streaming scheme applied to industrial monitoring}
    \label{applicative_case}
\end{figure*}

The availability of massive amounts of data, coupled with recent machine learning breakthroughs offers great potential in numerous domains and particularly for the industry. More specifically, in Industry 4.0 era, one major challenge is to exploit all information sources related to a system in order to perform data-driven monitoring for corrective and predictive maintenances. It comes with important scientific challenges: 
\begin{itemize}
    \item First, models must handle multimodal sources such as sensors measurements, textual maintenance reports, or machine images, among others appealing for multimodal learning.
    \item Targeted data are data streams that are heterogeneous by nature (time series, raw text, images, etc.) and by their acquisition frequency. Besides, these different streams are also unaligned, as the behaviour of a sensor at present time can be highly correlated with a maintenance report from several days or weeks in the past requiring the ability to cope with unaligned and long-range dependence.
    \item Finally, data history may be arbitrarily long, and input streams shall be processed in a streaming fashion at inference, as an industrial system may never stop (see Fig. \ref{applicative_case}).
\end{itemize}

These different challenges have been tackled in the literature but separately to the best of our knowledge. A large avenue of research exists in multimodal learning. See, for instance, ~\cite{Baltrusaitis} for a detailed survey. Recent approaches are often based on  Transformer models~\cite{vaswani2017attention} and their scaled dot-product attention module. Indeed, since their introduction, Transformer-based architectures have constituted breakthrough in many different Deep Learning fields, creating efficient contextualized encoders~\cite{devlin2019bert} and decoders~\cite{Radford2019LanguageMA}, or regularly beating SOTA benchmarks~\cite{dosovitskiy2021image, gulati2020conformer,DBLP:conf/icassp/DongXX18} on different unimodal data. Transformers models have been proposed to handle multimodal data~\cite{survey_transf} with interesting architectures such as Multimodal Transformer~\cite{Tsai2020a}, inferring unaligned temporal dependencies across modalities.

Due to their time and space complexity which is quadratic in the input sequence, multimodal transformer architectures fail to tackle the challenges of arbitrarily long inputs or streaming inference. Many approaches tried to alleviate this issue~\cite{tay2020efficient} by either using low-rank approximations of the self-attention matrix~\cite{katharopoulos2020transformers, choromanski2021rethinking}, adding some sparsity through selected or learned attention patterns~\cite{child2019generating, zaheer2021big,beltagy2020longformer, roy2020efficient}, or conveying information via a bounded memory \cite{dai2019transformerxl, rae2019compressive, martins2021inftyformer} decreasing the complexity up to a linear level.

At last, as mentioned before, for predictive maintenance, we have to consider stream data. To the best of our knowledge, handling stream data is mainly done for unimodal data. In particular, in the  Automatic Speech Recognition (ASR) domain, 
low-latency at test time is ensured, by chunking input sequences into smaller segments~\cite{wu2020streaming, tian2020synchronous, dong2019selfattention}. Notably, Emformer architecture~\cite{Shi2020} performs streaming ASR by updating a memory bank to convey information across segments. But this architecture is limited to unimodal sequences. \\

In this paper, we thus propose to tackle these three problems jointly and we propose a new architecture, named  StreaMulT, a Streaming Multimodal Transformer. 
Our global architecture, by combining crossmodal attention and block processing, can deal with heterogeneous and unaligned modalities. It enables to consider an arbitrarily long input multimodal data and to perform a streaming inference. \\

\textbf{Our contributions are threefold:}
\begin{itemize}
    \item First, we formally define a new applicative paradigm, in which one aims to solve a prediction task across time, from heterogeneous (by nature and acquisition frequency) multimodal sequential data and in a streaming fashion, hence handling arbitrarily long input data at both training and inference time. 
    \item We then propose StreaMulT, a Streaming Multimodal Transformer architecture to tackle this issue and deal with unaligned input streams.
    \item Due to the lack of a public dataset adapted to our task, we eventually propose to evaluate our model on CMU-MOSEI dataset with a multimodal sentiment analysis task, in order to compare StreaMulT performances with previous approaches. It includes both multimodal and unaligned streams. We show that our model can deal with arbitrarily long sequences without suffering from performance loss. We even improve the stat-of-the-art metrics on this dataset.  
\end{itemize}

We review existing strategies to tackle Fault diagnosis problem along with multimodal premises in section \ref{sec:relatedworks}. In section \ref{sec:SML}, we formalize our new paradigm. We then introduce our model, StreaMulT, in the section \ref{sec:proposedmodel}. At last, we conduct experiments on CMU-MOSEI dataset in Section \ref{experiments}.

\section{Related works}
\label{sec:relatedworks}

The pioneer series of three articles of Venkatasubramanian et al. \cite{VENKATASUBRAMANIAN} is one of the first works to list and categorize the different methods of fault diagnosis; and therefore constitutes the starting point of our review. 
This series classifies fault diagnosis approaches depending on both the a priori knowledge one has on eventual faults, along with how they would be expressed through the acquired data of the system (\textit{i.e.} fault symptoms). 
Strategies using this a priori knowledge by representing the system by a physical model are called "model-based" and divided between qualitative and quantitative depending on the mathematical representations that are used. 
On the other side, the approaches only relying on the history of acquired data are called "data-based". 
While model-based methods can be well suited when one has a nice a priori understanding of physical laws governing the system, they become less relevant otherwise. 
Thus, when the considered system reaches a certain level of complexity, inter-components interactions can less easily be modelled. 
To address this, data-based approaches provide a viable alternative : the designed model aims to learn these components dependencies from the data history. 
We mostly focus on data-based works in our review, and more precisely on Machine Learning (ML) ones. 

\subsection{Machine Learning for Fault Diagnosis}
Several reviews (cited thereafter) list the different ML architectures designed for tackling the fault diagnosis problem. Some of these reviews adopt an industrial-domain-specific position : while \cite{chemical} expose fault diagnosis methods that have been used for chemical process systems, \cite{bearing} focus on bearing faults, whereas \cite{conditioning} only consider residential air conditioning systems. 
These studies mainly motivate their approach by the consequences of fault occurrences in their relative fields, such as the over-consumption of electricity and the induced economic costs \cite{conditioning}. 
Besides, the methods listed in these reviews are presented as relevant for dealing with data relative to these applicative fields. 
Therefore, \cite{bearing} essentially consider vibration and stator current data, as contained in the Paderborn dataset\footnote{Available online: https://mb.uni-Paderborn.de/kat/forschung/datacenter/bearing-datacenter}; whereas \cite{conditioning} rather present models calibrated for thermostat and humidity data. 
By contrast, other works adopt a more methodological position regarding their reviews of state-of-the-art algorithms for addressing fault diagnosis \cite{Palade}. 
Most recent ones \cite{Angelopoulos, reis, Li2018DeepLD} motivate their work by the appearance of new practical challenges induced by the arrival of Industry 4.0 era, such as notably the ability to handle massive and multi-sources data with a short-time response.
These reviews qualify ML methods as more effective when fault profiles are complex. 
Thus, \cite{bearing} mention the limits of model-based approaches for the early detection of faults, due to symptoms that are untraceable by this kind of models. 
They also point out their difficulty to disentangle the simultaneous occurrences of different faults. \\
Although some articles only consider fault detection \cite{detection, snapshot}, the vast majority also consider fault isolation and identification (note that Angelopoulos et al. \cite{Angelopoulos} sometimes use the word "diagnosis" to evoke fault detection though). 
However as emphasized by \cite{reis}, in practice two methodologies co-exist. 
On the one hand, Statistical process control (SPC) community sequentially processes fault detection and fault isolation and identification. 
On the other hand, ML community often processes these two tasks in a simultaneous fashion, in the form of a $(C+1)$-classes classification, decomposed into one class of normal functioning mode and $C$ distinct faulty functioning modes. \\
As presented in \cite{Li2018DeepLD}, ML models used for fault diagnosis are generally composed of a feature-extraction module and a diagnosis module. 
In that configuration, the former feeds the latter relevant elements computed from raw data. Some feature-extraction modules focus on time domain to catch and characterize information contained within time series acquired from the system sensors, using for instance neural networks \cite{zarei}.
It is also common to use signal processing tools in order to exploit features from the time series in the frequency domain.
\cite{fourier} and \cite{laplace} thus respectively use Fourier and Laplace transforms to this purpose.
Finally, other approaches  chose to work in the time-frequency domain, through the usage of wavelet transforms for instance \cite{wavelet}. 
The choice of feature-extraction module is strongly influenced by the structure of input data and the subsidiary task, therefore by the a priori knowledge of its designer. \\
The very diagnosis module is then composed of:
\begin{itemize}
	\item either a first detection submodule aiming to perform fault monitoring, followed by a second classification submodule performing fault isolation and identification
	\item either a unique classification module carrying out simultaneously both fault detection and fault isolation and identification
\end{itemize}

In a supervised setting, the unique classification module fed with extracted features is free to use any ML model: Support Vector Machine \cite{konar}, Random Forest \cite{randomforest}, shallow neural networks \cite{shallownn}, recurrent neural networks \cite{yam}, and so on.
This scheme of performing simultaneously fault detection and classification has however been sometimes criticized \cite{reis}, as it might lead to practical issues:
\begin{itemize}
	\item fault occurrences that might lead to failures and dreaded events are often scarce in real datasets. This results in an imbalanced dataset problem, exacerbated the more faulty classes one considers.
	\item For this kind of tasks, a prediction error will have the same weight during the learning phase, regardless of which misclassification has been made. However, depending on the system criticality, one would like to put a lot more emphasis on the fault detection rather than on its proper identification. 
\end{itemize}
To cope with these issues, a prior monitoring task can be realised using anomaly detection methods \cite{anomaly}.
Similarly to the architectures designed in SPC community's works, these semi-supervised methods model the normal functioning mode of the system during the learning stage, a
nd classify as fault the datapoints which deviate significatively from this model's prediction at test time. 
These approaches are more robust to imbalanced datasets and can then be coupled with a classification model to perform the isolation and identification task. 
Lastly, if the normal functioning mode conditions are unknown (\textit{i.e.} in an unsupervised setting), it is also possible to design the diagnosis module by using clustering approaches. That is what are doing \cite{clustering}, comparing the performances of different unsupervised algorithms: Gaussian mixture, hierarchical clustering and K-means models. 

\subsection{Deep Learning}
In a similar way to model-based methods, over the last few years classical ML approaches faced some limitations induced by growing complexity of industrial system data. 
Indeed, as described in \cite{bearing, Li2018DeepLD, peng}, classical features extraction models based on a certain a priori knowledge on input data structure, may no longer be efficient to perform a correct fault diagnosis. 
To answer these challenges, Deep Learning (DL) models are designed, as they integrate a representation learning part in the first layers. 
This part aims to automatically extract the most salient features for a subsidiary task (here the fault diagnosis), with no - or few - a priori knowledge on input data structure required \cite{representation, deeplearning}. 
Thus, numerous articles have shown the superiority of DL models over classical ML ones for fault diagnosis, using as representation learning algorithms either discriminative models (CNN \cite{cnn1,cnn2,cnn3}, deep RNN \cite{lstm1,lstm2}, Transformers \cite{transformerdiag}, etc.) or generative models (PGM \cite{dbn1, dbn2}, autoencoders \cite{ae1,ae2,ae3}, GANs \cite{gan1, gan2}).

\begin{figure*}
    \centering
    \includegraphics[width = \textwidth]{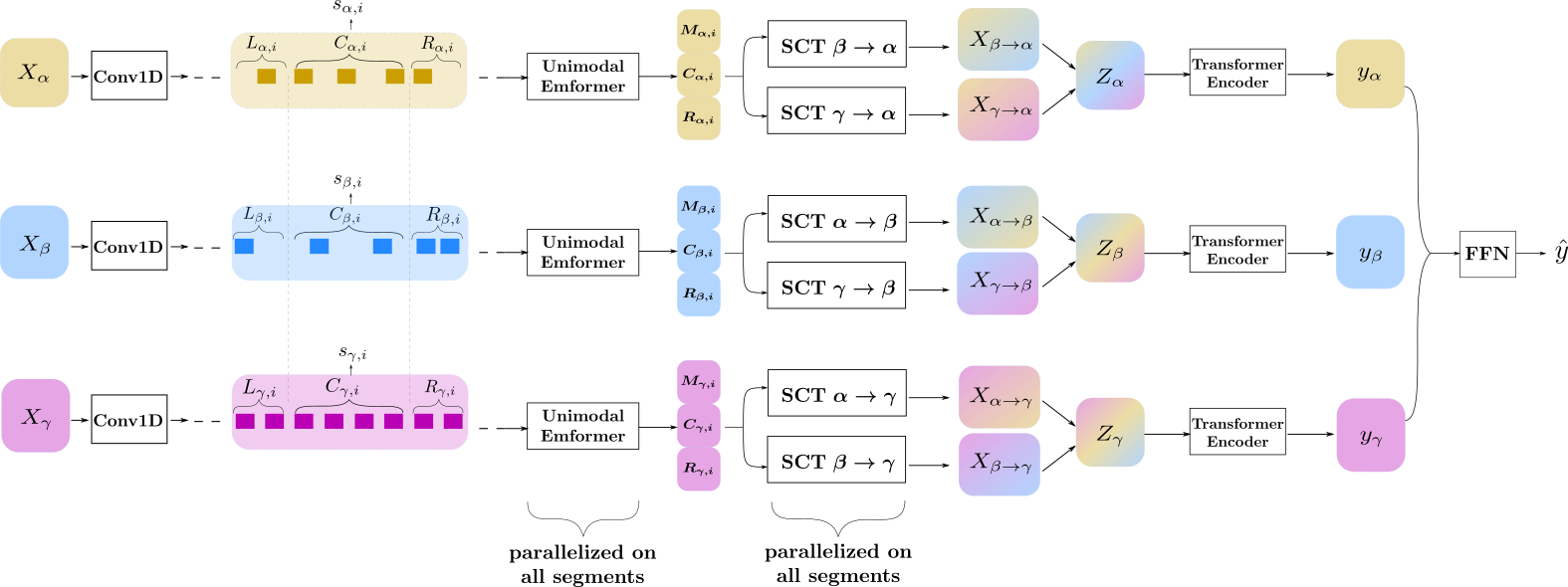}
    \caption{Streaming Multimodal Transformer architecture. SCT stands for Streaming Crossmodal Transformer. Different colors represent heterogeneity nature of different modalities, and shadings represent crossmodal features.}
    \label{fig:fullarchitecture}
\end{figure*}

\subsection{Fault diagnosis using multimodal data}
The complexity of industrial systems, along with the relative acquired datasets, reaches nowadays a new level, with sensors producing multimodal data. 
While some previous works tackled the challenge of fault diagnosis from thermal images \cite{therm1, therm2, therm3}, x-ray data \cite{xray}, photographs \cite{photos1, photos2} or textual maintenance reports \cite{logs1, logs2}, the application of such models to multimodal data (\textit{i.e.} of heterogeneous natures) is still in its infancy.
Most previous works addressing the fault diagnosis task and mentioning "multimodal" data actually refer to the different functioning modes of the considered system (such as an air conditioner functioning in eco-mode or in normal mode) \cite{sipple}. 
For \cite{curvature}, the word "multimodal" refers to the different orders of derivatives of the input time series. 
To the best of our knowledge, only two articles properly consider multimodal data (as of heterogeneous natures) in a perspective of industrial maintenance. 
\cite{tauheed} fuse numerical time series of vibration signals with thermal images in order to improve classification performances. They use a classical ML approach, with an Hilbert tranform module for feature extraction and a concatenation module for data fusion. 
Unfortunately, their dataset has not been made publicly available, which prevents the community to compare one's work to theirs. \\

We now introduce the new framework that we consider in this article. 

\section{Multimodal learning in streaming}
\label{sec:SML}
In this section, we define the challenging problem our method tackles.
For purposes of clarity, we consider three modalities, denoted by $\alpha, \beta, \gamma$. This case can be extended to any number of modalities without loss of generality. We consider 3 time series $(X_\alpha, X_\beta, X_\gamma)$ from different modalities (e.g. text, image, sound, numerical, etc.) as our input data. Each series is indexed by time, according to its own acquisition times and lies in its own definition space. Hence for the modality $\alpha$,
\begin{equation*}
    X_\alpha:= \left(X_\alpha(t)\right)_{t \in \mathcal{T}_\alpha} \text{ and } \hspace{.1em} \forall t \in \mathcal{T}_\alpha, \hspace{.5 em} X_\alpha(t) \in \mathbb{R}^{d_\alpha}
\end{equation*} 
where $\mathcal{T}_\alpha$ and $d_\alpha$ are respectively the countable set containing acquisition times of modality $\alpha$ and its associated feature dimension. \\
Our objective is to enable some prediction tasks (regression or classification) across time.
Let $\mathcal{X}:=  \left\{\left[X(s)\right]_{s\leq t}, t \in \mathbb{R}\right\}$ be the input space,
where $\left[X(s)\right]_{s\leq t}$ are data of all modalities acquired before time step $t$.
Formally, given a labeling space $\mathcal{Y}$ that is common to the different modalities, we try to find the optimal prediction function $h^* \colon \mathcal{X} \mapsto \mathcal{Y}$ minimizing a loss $L$ on some hypothesis space $\mathcal{H}$: 
\begin{equation*}
    h^*= \argmin_{h \in \mathcal{H}} L(h)
\end{equation*}
with \hspace{1em} $L(h):=\frac{1}{|\mathcal{T}_y|}\sum_{t\in\mathcal{T}_y}l\left(h\left(\left[X(s)\right]_{s\leq t}\right), y_t\right)$

where $l$ is a score function and $\mathcal{T}_y$ is the ground truth time steps, whose definition depends on the subsidiary task.
For instance, in the industrial monitoring application, \\ $\mathcal{T}_y := \mathcal{T}_\alpha \cup \mathcal{T}_\beta \cup \mathcal{T}_\gamma$ as the objective is to detect a fault at any time. In that example, different modalities $(\alpha, \beta, \gamma)$ can respectively correspond to numerical sensors measurements, system photographs, and textual maintenance reports. \\
However, if we now consider a sequence-classification task on an arbitrarily long corpus (keeping past sentences as input), then for a corpus of $s$ multimodal sentences, the associated ground truth time steps are the last acquisition time steps of each sentence: 
\begin{equation*}
    \mathcal{T}_y = \left\{\max_{ \mathcal{T}_\alpha^j\cup\mathcal{T}_\beta^j\cup\mathcal{T}_\gamma^j} t, \hspace{.5 em}  1\leq j \leq s \right\}
\end{equation*}
where $j$ is the sentence index. In this case, the different modalities $(\alpha, \beta, \gamma)$ can respectively correspond to visual frames of the speaker's face, sound recording of the sentences, and textual transcript of the sentences.



To the best of our knowledge, this paradigm has never been introduced as such. In the following section we introduce a new architecture to address our objective.

\section{Proposed model}
\label{sec:proposedmodel}
We propose StreaMulT, a Streaming Multimodal Transformer architecture, taking advantages of both a crossmodal attention mechanism and a block processing approach to tackle the different challenges of this framework. The architecture is illustrated in Fig. \ref{fig:fullarchitecture}. Finally, we optimize the training scheme of the model to lower space complexity, training time and enabling inference short-time response at the same time. 
\subsection{Crossmodal Transformer and Block processing reviews}
\label{ssec:crossmodaltransformer}
Crossmodal Attention module, as defined in~\cite{Tsai2020a}, deals with heterogeneity gap of multimodal inputs~\cite{8715409} by expressing a target modality $\alpha$ with raw features from a source modality $\beta$. Formally, considering our input sequences $X_\alpha$\ and $X_\beta$ from modalities $\alpha$ and $\beta$, the crossmodal attention for $X_\alpha$ attending to $X_\beta$ , denoted $X_{\beta \rightarrow \alpha}$ is computed as:   
\begin{align*}
X_{\beta \rightarrow \alpha} :&= \text{softmax}\left(\frac{Q_\alpha K_\beta^T}{\sqrt{d_k}}\right)V_\beta \\
&=\text{softmax}\left(\frac{X_\alpha W_{Q_\alpha} W_{K_\beta}^TX_\beta^T}{\sqrt{d_k}}\right)X_\beta W_{V_\beta}
\end{align*}
with $(Q_\alpha)$ the query matrix for modality $\alpha$, $K_\beta, V_\beta$ the key and value matrices for modality $\beta$ and $W_{Q_\alpha}, W_{K_\beta}, W_{V_\beta}$   being learned weights. This scaled dot-product attention, inspired by original Transformer self-attention~\cite{vaswani2017attention}, models long-term dependencies through its matrix product. Hence it succeeds in handling unaligned multimodal data in the same way \cite{Tsai2020a}.
\\

\begin{figure}
    \centering
    \includegraphics[width = .45\textwidth]{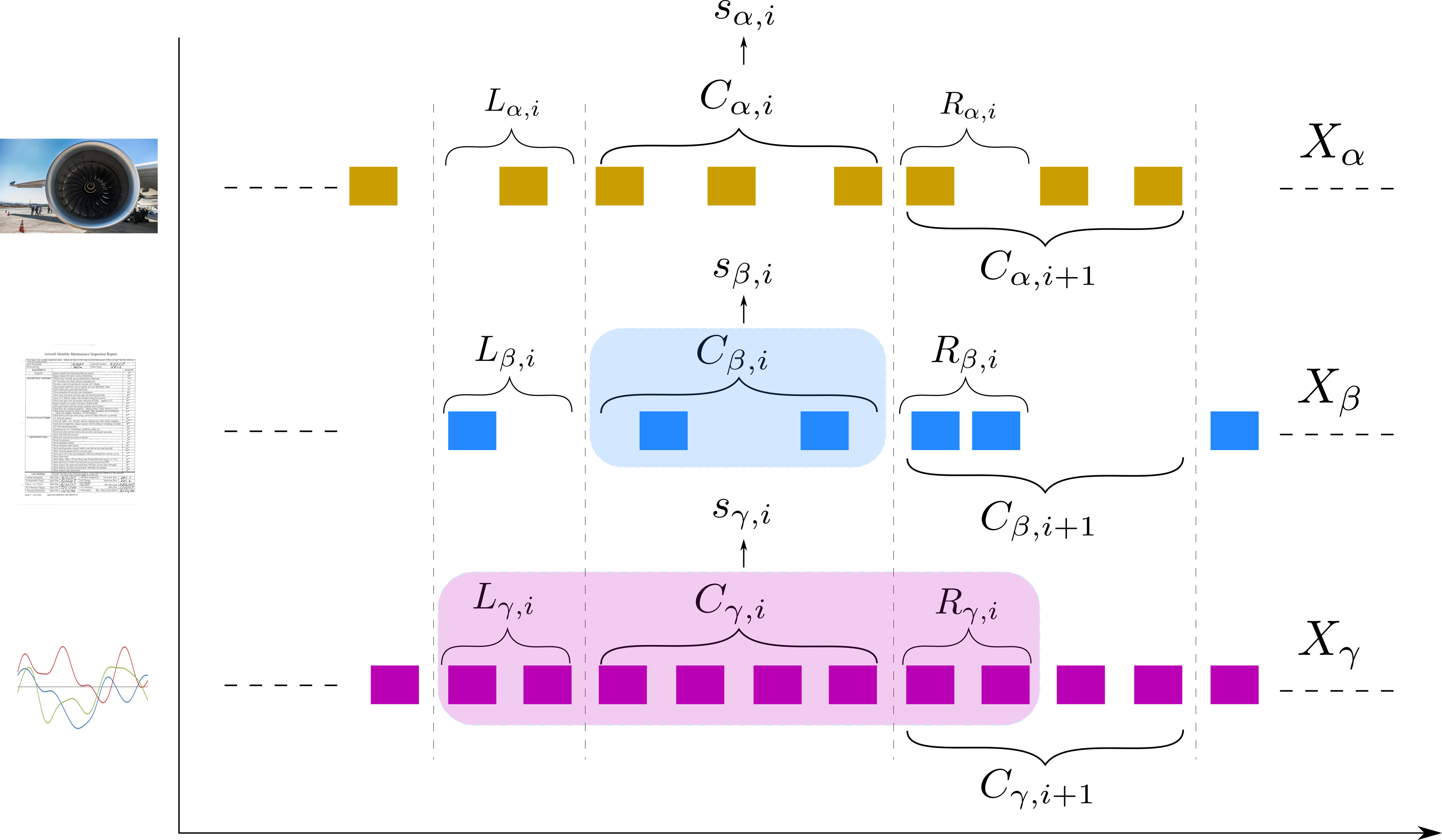}
    \caption{Block processing for Multimodal learning in a streaming scheme. For modality $\alpha$: $X_\alpha, C_{\alpha,i}, L_{\alpha, i}$ and $R_{\alpha, i}$ respectively correspond to the full input sequence, the initial $i$-th block, and the left and right contexts associated to this block to form the contextual $i$-th segment. $s_{\alpha,i}$ corresponds to the mean of current segment $C_{\alpha,i}$. Blue area represents an initial block for modality $\beta$ while the pink one represents a contextual segment for modality $\gamma$.}
    \label{fig:blockprocessing}
\end{figure}

However, due to the arbitrarily long size of input sequences in our setting, Multimodal Transformer architecture faces two main issues. Training is intractable due to its quadratic complexity, and inference cannot be done in a streaming way, as the vanilla model needs the whole sequence as input to compute the matrix product. To alleviate this, we use block processing method, chunking input sequences into non-overlapping smaller segments $(C_i)_{i\geq0}$ (see Fig.~\ref{fig:blockprocessing}). We then compute attention on these segments and hence reduce complexity during the cross-modal attention computation. Extending the block processing method to input data with heterogeneous sampling rates, we define hard segment bounds with respect to the temporal axis, hence producing shared segments across modalities. To prevent boundary effect, left and right context blocks are concatenated with initial blocks to form contextual segments $X_i = \left[L_i : C_i : R_i \right]$.

An Augmented-Memory Transformer ~\cite{wu2020streaming} approach then encodes segments information, by learning and storing a memory bank to convey information through time. Caching values from previous segments for left-context representations instead of recomputing attention makes the approach more efficient. \\

Considering a contextual segment $X_i = [L_i:C_i:R_i]$ and a memory bank $M_i = [m_1, \ldots, m_{i-1}]$ containing compressed information from previous segments, the output $X_i^{n+1}$ of the $n$-th layer is computed as:
\begin{small}
\begin{align}
    \hat{X}_i^{n} &= \text{LN}(X_i^n)  \\
    K_i^n &= W_k[M_i^n, \hat{X}_i^n] \\
    V_i^n &= W_V[M_i^n,\hat{X}_i^n] \\
    Q_i^n &= W_Q\hat{X}_i^n \\
    [Z_{L,i}^n:Z_{C,i}^n:Z_{R,i}^n] :&= \text{Attn}\left(Q_i^n,K_i^n,V_i^n\right) + X_i^n \\
    \hat{X}_i^{n+1} &= \text{FFN}\left(\text{LN}\left([Z_{L,i}^n:Z_{C,i}^n:Z_{R,i}^n]\right)\right) \\
    X_i^{n+1} &= \text{LN}\left(\hat{X}_i^{n+1} + [Z_{L,i}^n:Z_{C,i}^n:Z_{R,i}^n]\right) \\
    m_i^n &= \text{Attn}\left(W_Qs_i^n, K_i^n, V_i^n\right)
\end{align}
\end{small}
\noindent where $s_i^n$ is the mean of $C_i^n$ and LN, FFN, Attn respectively correspond to Layer Normalization, Feed-Forward and scaled dot-product Attention layers. After passing through all $N$ layers, outputs corresponding to left and right contexts are discarded to keep only center segments representations $(C_i^N)_{i\geq0}$. \\

\subsection{Putting together with Memory bank}
\label{ssec:membank}

\begin{figure}
    \centering
    \includegraphics[width = .45\textwidth]{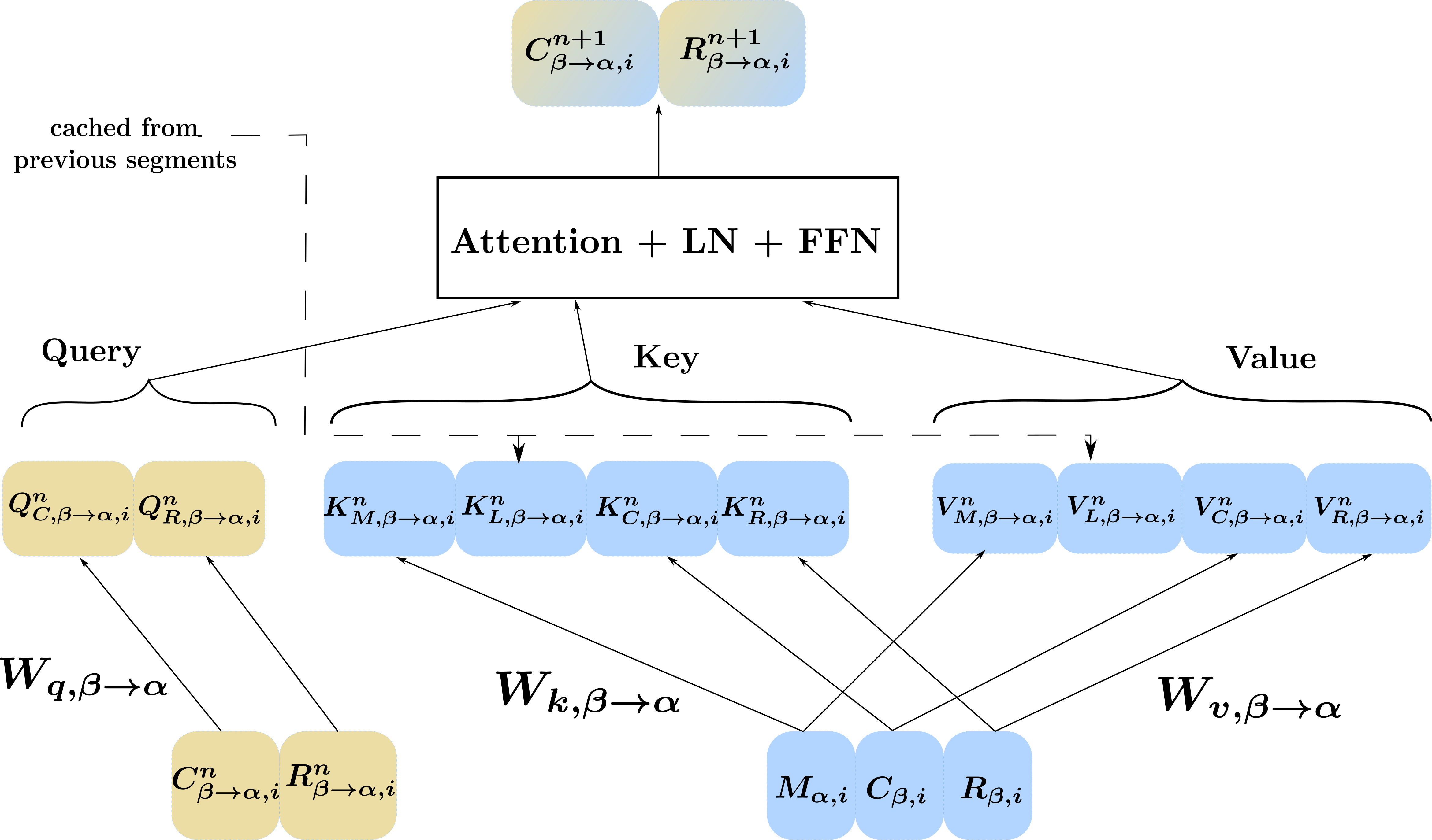}
    \caption{Streaming Crossmodal Transformer module \vspace{7.25 em} \phantom{a}}
    \label{fig:SCT}
\end{figure}

Our global end-to-end architecture combines benefits from Block processing and Crossmodal Attention. The architecture is illustrated in Fig.~\ref{fig:fullarchitecture}. We describe here the processing of the modality $\alpha$. \\
$X_\alpha$ is first passed through a 1D convolutional layer aiming to model some local temporal structure, and map all modalities to a common feature dimension $d$.
 Segment bounds are then fixed, and following block processing approach, every contextual segments $X_{\alpha, i}$ are processed in a parallel way. They are first given to a modality-specific Emformer to initialize its own modality memory bank $M_\alpha$. Then, each source modality / target modality ($\beta$ / $\alpha$) pair is processed by its own Streaming Crossmodal Transformer (SCT) module. Specifically, each segment from the target modality $X_{\alpha, i} = \left[L_{\alpha,i}:C_{\alpha,i}:R_{\alpha,i}\right]$ is expressed using the same temporal segment from the source modality $X_{\beta, i}$ along with the source modality memory bank $M_{\beta, i}$. For each layer $n$:
\begin{small}
\begin{align*}
    &\left[\hat{C}_{\alpha, i}^n, \hat{R}_{\alpha, i}^n\right]  = \text{LN}(\left[C_{\alpha, i}^n, R_{\alpha, i}^n\right])  \\
    &\left[\hat{C}_{\beta, i}^n, \hat{R}_{\beta, i}^n\right]  = \text{LN}(\left[C_{\beta, i}^n, R_{\beta, i}^n\right])  \\
    &K_{\beta,i}^n = \left[K_{M,\beta\rightarrow\alpha,i}^n, K_{L,\beta \rightarrow \alpha,i}^n, K_{C,\beta\rightarrow\alpha,i}^n,K_{R,\beta\rightarrow\alpha,i}^n\right] \\
    &V_{\beta,i}^n = \left[V_{M,\beta\rightarrow\alpha,i}^n, V_{L,\beta \rightarrow \alpha,i}^n, V_{C,\beta\rightarrow\alpha,i}^n,V_{R,\beta\rightarrow\alpha,i}^n\right] \\
    &Z_{C, \beta \rightarrow \alpha, i}^n = Attn(Q_{C,\beta\rightarrow\alpha,i}^n, K_{\beta, i}^n, V_{\beta, i}^n) + C_{\beta \rightarrow \alpha,i}^n \\
    &Z_{R, \beta \rightarrow \alpha, i}^n = Attn(Q_{R,\beta\rightarrow\alpha,i}^n, K_{\beta, i}^n, V_{\beta, i}^n) + R_{\beta \rightarrow \alpha,i}^n \\
    &\left[\hat{C}_{\alpha, i}^{n+1}, \hat{R}_{\alpha, i}^{n+1}\right]  = \text{FFN(LN}([Z_{C, \beta \rightarrow \alpha, i}^n, Z_{R, \beta \rightarrow \alpha, i}^n])) \\
    &\left[C_{\alpha, i}^{n+1}, R_{\alpha, i}^{n+1}\right]  = \text{LN}(\left[\hat{C}_{\alpha, i}^{n+1}, \hat{R}_{\alpha, i}^{n+1}\right] + [Z_{C, \beta \rightarrow \alpha, i}^n, Z_{R, \beta \rightarrow \alpha, i}^n])
\end{align*}
\end{small}

\noindent where,
\begin{footnotesize}
\begin{align}
    &\left[K_{M,\beta\rightarrow\alpha,i}^n,K_{C,\beta\rightarrow\alpha,i}^n,K_{R,\beta\rightarrow\alpha,i}^n\right]=W_{k,\beta\rightarrow\alpha}\left[M_{\beta,i},\hat{C}_{\beta,i}^n,\hat{R}_{\beta,i}^n\right] \\
    &\left[V_{M,\beta\rightarrow\alpha,i}^n,V_{C,\beta\rightarrow\alpha,i}^n,V_{R,\beta\rightarrow\alpha,i}^n\right]=W_{v,\beta\rightarrow\alpha}\left[M_{\beta,i},\hat{C}_{\beta,i}^n,\hat{R}_{\beta,i}^n\right] \\
    &\left[Q_{C,\beta\rightarrow\alpha,i}^n,Q_{R,\beta\rightarrow\alpha,i}^n\right]=W_{q,\beta\rightarrow\alpha}\left[C_{\beta\rightarrow\alpha,i}^n,R_{\beta\rightarrow\alpha,i}^n\right]
\end{align}
\end{footnotesize}

\noindent and $\left(K_{L,\beta \rightarrow \alpha,i}^n, V_{L,\beta \rightarrow \alpha,i}^n\right)$ are the key and value copies (cached) corresponding to previous segments, up to left context size. This module is illustrated in Fig. \ref{fig:SCT}.

\begin{figure*}[h]
    \centering
    \includegraphics[width = \textwidth]{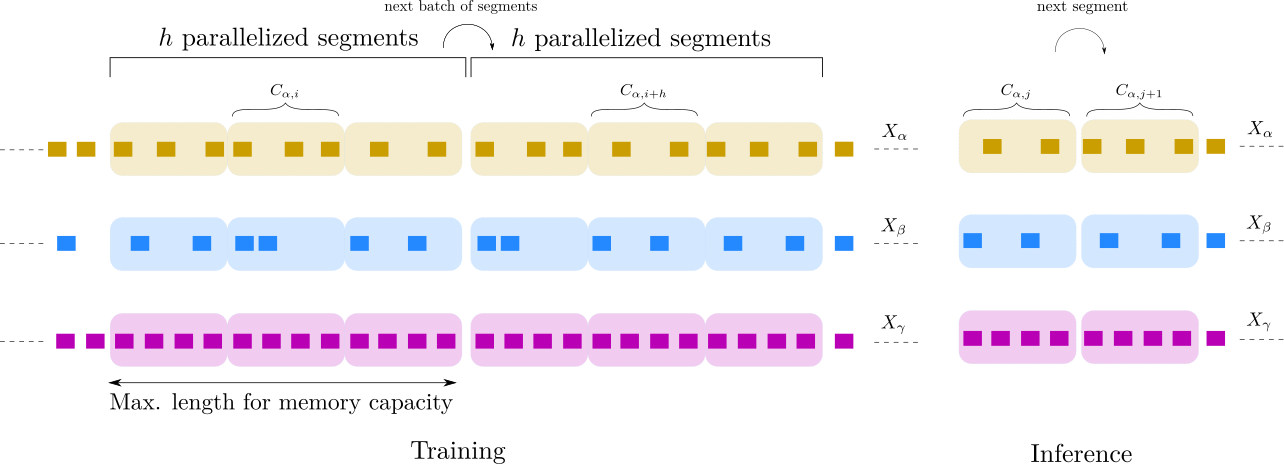}
    \caption{Flexible scheme. At training time (left), subsequences of $h$ consecutive segments are created to parallelize crossmodal attention operations. At inference (right), one can still process segments one by one to obtain a short-time response.}
    \label{fig:training}
\end{figure*}

    \begin{table*}[!b]
        \caption{Time complexity, Space complexity and number of sequential operations for different layer types.}
        \begin{small}
        \begin{center}
        \begin{tabular}{|c|c c c|} 
           \hline
            Layer type & Time Complexity & Space Complexity & Sequential Operations \\
            \hline
            Self-Attention & $O(n^2.d)$ & $O(n^2 + n.d)$ & $O(1)$ \\
            Cross-Attention & $O(n_\alpha.n_\beta.d)$ & $O(n_\alpha.n_\beta + n_\alpha.d + n_\beta.d)$ & $O(1)$ \\
            \begin{tabular}{@{}c@{}} Streaming Cross-Attention\\   (regular training scheme) \end{tabular} & $O(n_\alpha.n_\beta.d)$ & $O(n_\alpha.n_\beta + n_\alpha.d + n_\beta.d)$ & $O(1)$ \\
            \begin{tabular}{@{}c@{}} Streaming Cross-Attention \\ (flexible training scheme) \end{tabular} & $O(n_\alpha.h.C_\beta.d)$ & $O(h^2.C_\alpha.C_\beta + h.C_\alpha.d + h.C_\beta.d)$ & $O\left(\frac{n_\alpha}{hC_\alpha}\right)$ \\
            \hline
        \end{tabular} 
        \end{center}
        \label{tab:complex}
        \end{small}
    \end{table*}

After the last layer $N$, right contexts representations $(R_{\beta \rightarrow \alpha, i}^D)_{i > 0}$ are discarded. $(C_{\beta \rightarrow \alpha, i}^D)_{i > 0}$ are concatenated to form the final crossmodal representation $X_{\beta\rightarrow\alpha}$. We then concatenate along the feature dimension all crossmodal outputs corresponding to the same target modality $\alpha$ in a vector $Z_\alpha := \begin{pmatrix} X_{\beta\rightarrow\alpha} \\X_{\gamma\rightarrow\alpha} \end{pmatrix}$, that is given as input to a Transformer Encoder exploiting sequential nature of data, to produce modality output $y_\alpha$. All modality outputs are eventually concatenated and passed through a final fully-connected layer to output prediction $\hat{y}$. \\

\subsection{Training scheme: balancing space and time complexities}
\label{subsec:complex}
The main motivation to design StreaMulT architecture is to handle the arbitrarily long nature of considered multimodal input sequences. In that sense, the block processing mechanism we use aims to alleviate the quadratic complexity of cross-modal attention modules, similarly to several speech recognition works \cite{Shi2020, amtrf}. However, these applications focus on getting short-time response at inference to perform simultaneous speech translation or recognition and hence essentially differ from our framework. Indeed, to handle very long sequences we are at least as concerned about space complexity as time complexity. We thus cannot train our model in the same fashion as these approaches, that is by parallelizing on all input segments the cross-modal attention computation. This indeed still implies a quadratic space complexity to store cross-modal attention weights matrix. \\
To fulfill both space capacity and efficient training time constraints, we introduce a flexible training scheme. This is illustrated in Fig. \ref{fig:training}. More specifically, at training time we parallelize operations of Memory bank initialization and Streaming Crossmodal Transformer modules on subsequences of $h$ consecutives segments. $h$ is chosen in an empiric way, as the highest integer enabling one's memory capacity to run the model. This training scheme enables StreaMulT to run arbitrarily long sequences by only storing limited-size matrices, while still benefiting from simultaneous computations through parallelization. Note that we do not change the segment length but rather concatenate them in a single matrix product. This enables to keep short segments at inference and thus still work in a short-time response for streaming application. \\
Table \ref{tab:complex} derives the different time and space complexity classes for different types of layers, along with the number of sequential operations. Vanilla self-attention layers has a quadratic complexity both in time and in space, which is problematic for handling long sequences. Similarly, crossmodal attention, as defined in \cite{Tsai2020a} also has a quadratic complexity in the sequence length. More precisely, the complexity class depends of the product of the two modalities lengths $n_\alpha,.n_\beta$, as they can differ.\\
Streaming Crossmodal Attention modules trained in regular fashion for blocks processing (as in \cite{Shi2020}) have the same space and time complexity classes, which make them untractable for arbitrarily long sequences. This indeed requires to compute the matrix product of $Q_\alpha\in\mathbb{R}^{d_{q_\alpha},d}$ and $K_\beta\in\mathbb{R}^{d_{k_\beta},d}$, with $d_{q_\alpha} = n_{\text{seg}}.(R_\alpha+C_\alpha+1)$ and $d_{k_\beta} = n_{\text{seg}}.(R_\beta+C_\beta+l_{\text{mem}})$. $n_\text{seg}$ is the number of segments of the input sequence, $R_\alpha$ and $R_\beta$ corresponds to the length of right contexts for modalities $\alpha, \beta$, and $C_\alpha$ and $C_\beta$ to the length of their central segments. Last, $l_{\text{mem}}$ corresponds to the length of a memory cell. We suppose that $R$ and $l_{\text{mem}}$ are negligible before $C$, and noting that $n_{\text{seg}} = \frac{n_\alpha}{C_\alpha} = \frac{n_\beta}{C_\beta}$, one obtains the results mentioned above. \\
If we train this layer in the flexible scheme as described in section \ref{subsec:complex}, for each subsection of $h$ consecutive segments we need to handle the product of matrices $Q_\alpha\in\mathbb{R}^{d_{q_\alpha},d}$ and $K_\beta\in\mathbb{R}^{d_{k_\beta},d}$, with now $d_{q_\alpha} = h.(R_\alpha+C_\alpha+1)$ and $d_{k_\beta} = h.(R_\beta+C_\beta+l_{\text{mem}})$, which has a time complexity class of $O(h^2.C_\alpha.C_\beta.d)$. As mentioned in the third column, to process the whole sequence we need to perform $\frac{n_\alpha}{hC_\alpha}$ sequential operations, which also derives the whole time complexity class. Note that the space complexity now only depends on $h, C$ and $d$, as we only need to store a subsequence at a time. \\
At inference, one can thus choose $h=1$ to process the input sequence in streaming, enabling a short-time response with time and space complexity classes being respectively $O(C_\alpha.C_\beta.d)$ (for one segment) and $O(C_\alpha.C_\beta+C_\alpha.d + C_\beta.d)$.

\section{Experiments and results}
\label{experiments}
\subsection{Dataset and setups}
Despite having a public dataset compatible with the Streaming Multimodal Learning challenge, involving long, heterogeneous and unaligned input sequences, we conduct experiments on CMU-MOSEI dataset~\cite{bagher-zadeh-etal-2018-multimodal}, to empirically evaluate the StreaMulT architecture and compare it with existing approaches handling sequential unaligned multimodal data. CMU-MOSEI dataset consists of 23,454 movie review video clips on YouTube, from which are extracted audio and video features using Facet (based on CERT~\cite{5771414}) and COVAREP~\cite{inproceedings}. Textual features are also extracted from words transcripts, using Glove~\cite{pennington-etal-2014-glove} pretrained embeddings. This produces an unaligned version of the dataset, which is used to create a word-aligned version, using P2FA algorithm~\cite{Yuan2008SpeakerIO}. All aligned sentences are padded to a fixed length of 50 time steps. \\
The related task aims to perform sentiment analysis on these clips, labeled by human annotators with a sentiment score from -3 to 3. As in~\cite{Tsai2020a} and previous works, we evaluate model performances using various metrics: 7-class-accuracy, binary accuracy (positive or negative statements), F1-Score, MAE and correlation between model's predictions and labels. \\
To highlight StreaMulT added value, we conduct experiments in different settings. We first consider input video clips as our whole input sequences, and observe StreaMulT performances when dividing these clips into smaller segments. As we need to define hard segment temporal bounds, which are not given in the unaligned version of CMU-MOSEI, we conduct this experiment with the aligned version of the dataset. For StreaMulT, we choose to divide the input sentences into 5 segments of length 10. 


We compared StreaMulT performances with Multimodal Transformer (MulT) and other models addressing Multimodal Sentiment Analysis challenge, among which the recent SOTA methods \cite{self_mm, unknown}. We strongly emphasize that the added value of StreaMulT is its ability to deal with arbitrarily long unaligned multimodal inputs, and that it does not intend to address Multimodal Sentiment Analysis specific task. Hence at first we only reported Multimodal Transformer metrics scores given in~\cite{Tsai2020a} for a fair comparison, as both approaches use GloVe embeddings for text modalities whereas most recent works \cite{self_mm, unknown} use BERT embeddings. We also used the available official code\footnote{https://github.com/yaohungt/Multimodal-Transformer} for Multimodal Transformer architecture to run the experiments, with hyperparameters given in~\cite{Tsai2020a}. We could not reproduce the results shown in the paper, hence we present the results we obtained, that are not as good as the given ones. All scores from our experiments are averaged on 5 runs. The corresponding results are represented in the upper part of following Table \ref{tab:results}. This shows that our architecture globally reproduces the results of Multimodal Transformer (even performs a little bit better on some metrics), which highlights the availability of its memory bank to properly convey salient information through time, as StreaMulT receptive field only attends to segments of length 10, while MulT attends to whole sequence of length 50. \\
We then decided to use contextual pretrained embedding layers for textual modality, namely BERT \cite{devlin2019bert} and BART \cite{BART}. The corresponding results are described in the lower part of Table \ref{tab:results}, with a significant improvement in all metrics, StreaMult-BART achieving now the best results on the aligned version of CMU-MOSEI dataset.

    \begin{table}[t]
        \caption{Results on CMU-MOSEI aligned. Best results are marked in bold.\ddag: results from \cite{Tsai2020a}. *: Own implementation or reproduced from official code with provided hyper-parameters.}
        \begin{footnotesize}
        \centering
        \begin{tabular}{|c||c c c c c|} 
           \hline
            Metric &$\text{MAE}^l$ & $\text{Corr}^h$ & $\text{Acc}_7^h$ & $\text{Acc}_2^h$ & $\text{F1}^h$\\
            \hline
            MulT$^\ddag$ &   $0.580$ &  ${0.703}$ & ${51.8}$ &  ${82.5}$ &  ${82.3}$  \\
            MulT$^*$ &0.615 & 0.666 & 49.32 & 81.05 & $81.42$  \\
            StreaMulT & $0.608$ & $0.671$ & $50,08$& $81.08$ & 81.01   \\

            \hline
            MulT-BERT$^*$ & 0.563 & 0.771 & 50.85 & 85.59 & 85.63 \\
            StreaMulT-BERT$^*$ & 0.551 & 0.764 & 52.04 & 85.46 & 85.56 \\
            MulT-BART$^*$ & 0.543 & 0.782 & 53.83 & 86.28 & 86.29 \\
            StreaMulT-BART$^*$ & \textbf{0.523} & \textbf{0.786} & \textbf{54.54} & \textbf{86.97} & \textbf{86.97}    \\  
            \hline
        \end{tabular} 
        \label{tab:results}
        \end{footnotesize}
    \end{table}

We then trained the Multimodal Transformer and StreaMulT architectures on unaligned version of CMU-MOSEI dataset and reported the results in Table \ref{tab:results_unaligned}.

\begin{table}[t]
    \caption{Results on CMU-MOSEI unaligned. Best results are marked in bold. \ddag: results from \cite{unknown}. $\natural$: results from \cite{Tsai2020a}.}
    \begin{scriptsize}
    \centering
    \begin{tabular}{|c||c c c c c|} 
       \hline
        Metric & $\text{MAE}^l$ & $\text{Corr}^h$ & $\text{Acc}_7^h$ & $\text{Acc}_2^h$ & $\text{F1}^h$  \\
        \hline
        TFN$^\ddag$ & 0.593 & 0.700 & 50.2 &  - /82.5 & - /82.1 \\
        LMF$^\ddag$ & 0.623 & 0.677 & 48.0 & - /82.0 & - /82.1 \\
        MFM$^\ddag$ &  0.568 & 0.717 & 51.3 & - /84.4 & - /84.3 \\
        ICCN$^\ddag$ & 0.565 & 0.713 & 51.6 & - /84.2 & - /84.2 \\
        MulT$^\natural$ & 0.591 & 0.694 & 50.7 & - /81.6 & - /81.6 \\
        MISA$^\ddag$  & 0.568 & 0.724 & - & 82.59/84.23 & 82.67/83.97\\
        MAG-BERT$^\ddag$ & 0.539 & 0.753 & - & 83.8/85.2 &  83.7/85.1 \\
        Self-MM$^\ddag$ & 0.530 & 0.765 & - & 82.81/85.17 &  82.53/85.30 \\
        MMIM$^\ddag$ & \textbf{0.526} & 0.772 & \textbf{54.24} & 82.24/85.97 & 82.66/85.94 \\
        \hline
        MulT-BERT & 0.544 & 0.776 & 52.86 & 82.85/85.95 & 83.18/85.97 \\
        MulT-BART & 0.532 & \textbf{0.792} & 54.17 & \textbf{84.11/86.9} &  \textbf{84.51/86.95} \\
        StreaMulT-BERT & 0.570 & 0.774 & 50.89 & 82.31/85.98 & 82.71/86.13 \\
        StreaMulT-BART & 0.531 & 0.778 & 53.89 & 83.30/86.35 & 83.74/86.39 \\
        \hline

    \end{tabular}
    \label{tab:results_unaligned}
    
    \end{scriptsize}
\end{table}

Once again, the usage of a contextual pretrained embedding layer significantly improves performances. The Multimodal Transformer architecture coupled with a BERT embedding layer now equals the performances of SOTA MMIM model on several metrics, questioning the real improvement on the Multimodal Sentiment Analysis task over the last three years. Besides, it emphasizes the power of language models, which is supported by the performances of MulT-BART, defining a new SOTA for several metrics on this dataset. \\

We finally simulated arbitrarily long sequences by concatenating all video clips related to the same speaker and considering these as inputs streams. In this setting, StreaMulT architecture successfully parallelizes its training along segments and handle long sequences at inference in a streaming way. On the other side, Multimodal Transformer faces memory issue. 

\begin{figure}[!t]
    \centering
    \includegraphics[width = .4\textwidth]{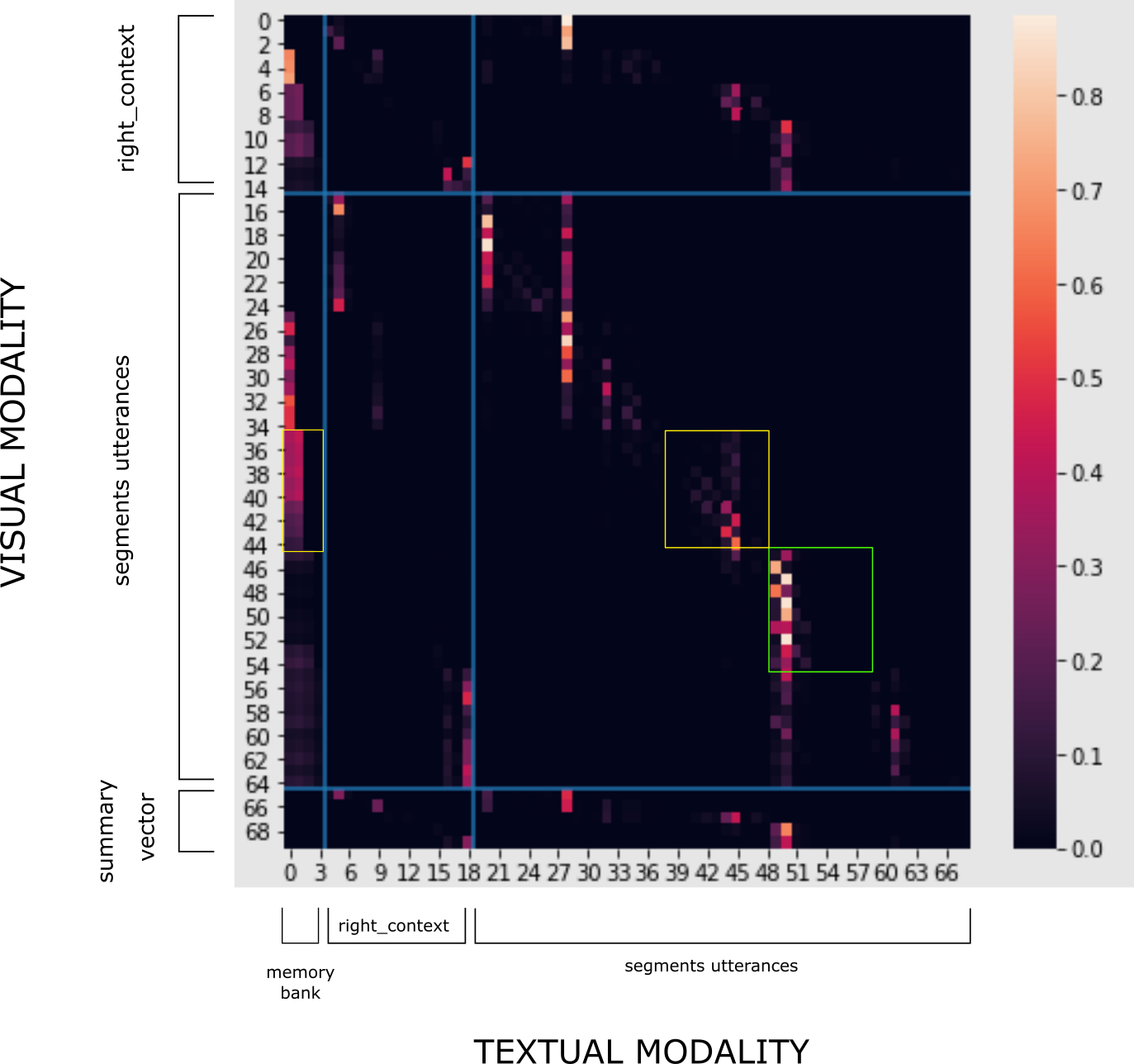}
    \caption{Heatmap of StreaMulT attention weights for the Visual/Textual crossmodal module. The sequence of length 50 is chunked into segments of length 10, with left and right contexts of respectively lengths 10 and 3.}
    \label{fig:attention_map}
\end{figure}

To qualitatively validate our architecture, we also plot the heatmap of the different attention weights of the model in the Fig. \ref{fig:attention_map}. \\
This plot represents the different attention weights of the Streaming Crossmodal Transformer related to the visual/textual modalities, for a multimodal sequence of length 50. For consistence with previous notations, we call $\alpha$ the visual modality and $\beta$ the textual modality. On the x-axis, the key matrix $K_\beta$ is organized as: \\ $\left[\text{memory bank; right contexts; segments utterances} \right]$. On the y-axis, the query matrix $Q_\alpha$ is organized as: \\
$\left[\text{right contexts; segments utterances; summary vectors}\right]$. Different blocks are delimited on Fig. \ref{fig:attention_map} by vertical and horizontal blue lines. \\

This figure first reminds us, as stated in \cite{Tsai2020a}, that language sequences are unaligned across modalities. This is indeed shown by the several activations on vertical lines (differing from a temporal monotonic diagonal line), corresponding to specific word embeddings correlated to many visual frames.\\
If some of these unalignments remain in the scope of the same temporal segment, as illustrated in the fourth segment by the green box, the access to the memory bank enables the model to attend beyond the current segment and to catch unalignments at longer range, as illustrated in the third segment by the yellow boxes. The yellow box on the right witnesses the unaligned dependencies within the third segment, while the left yellow box illustrates that some textual features of the past history activate the visual frames of the current segment. \\
These different behaviors show the ability of the StreaMulT architecture to adapt its strategy depending of the context, attending to unaligned data from past history via memory bank when necessary.

\section{Conclusion}

The proposed StreaMulT combines the power of crossmodal attention for multimodal representation  with the efficiency of block processing approach to process arbitrarily long sequence in a streaming fashion. That way, it addresses the newly introduced challenge of Multimodal Learning in Streaming, for which existent approaches struggle. Experiments conducted on CMU-MOSEI dataset showed promising results, with an improvement of the state of the art metrics and an ability to both handle arbitrarily long data at train time and process sequences in a streaming fashion at inference. Numerous applications of this paradigm such as Industrial Monitoring need an adapted dataset to compare related future works. \\

\newpage

\bibliographystyle{IEEEbib}
\bibliography{main}

\end{document}